\def\paperID{15328}
\def\confName{CVPR}
\def\confYear{2024}

\def\paperTitle{Consistency and Uncertainty: Identifying Unreliable Responses From Black-Box Vision-Language Models for Selective Visual Question Answering}

\def\authorBlock{
Zaid Khan \quad Yun Fu \\
\small{Northeastern University} \\
}

\newif\ifreview 
\newif\ifarxiv \newcommand{\arxiv}{\arxivtrue}
\newif\ifcamera 
\newif\ifrebuttal 

\arxiv 

\pdfoutput=1
\documentclass[10pt,twocolumn,letterpaper]{article}
\ifreview \usepackage[review]{cvpr} \fi
\ifarxiv \usepackage[pagenumbers]{cvpr} \fi
\ifrebuttal \usepackage[rebuttal]{cvpr} \fi
\ifcamera \usepackage{cvpr} \fi

\usepackage{graphicx}	
\usepackage{amsmath}	
\usepackage{amssymb}	
\usepackage{booktabs}
\usepackage{times}
\usepackage{microtype}
\usepackage{epsfig}
\usepackage[table,xcdraw,dvipsnames]{xcolor}
\usepackage{caption}
\usepackage{float}
\usepackage{placeins}
\usepackage{color, colortbl}
\usepackage{stfloats}
\usepackage{enumitem}
\usepackage{tabularx}
\usepackage{xstring}
\usepackage{multirow}
\usepackage{xspace}
\usepackage{url}
\usepackage{subcaption}
\usepackage{xcolor}
\usepackage[hang,flushmargin,symbol]{footmisc}
\usepackage{newtxmath}
\usepackage{wrapfig}
\usepackage{adjustbox}
\usepackage{multirow}
\usepackage{multicol}
\usepackage{colortbl}

\usepackage[ruled,vlined]{algorithm2e}
\usepackage{subcaption}

\ifcamera \usepackage[accsupp]{axessibility} \fi

% ...

\ifarxiv  \fi

\newcommand{\R}[1]{{%
    \textbf{%
        \ifstrequal{#1}{1}{\textcolor{red}{R#1}}{%
        \ifstrequal{#1}{2}{\textcolor{blue}{R#1}}{%
        \ifstrequal{#1}{3}{\textcolor{magenta}{R#1}}{%
        \ifstrequal{#1}{4}{\textcolor{teal}{R#1}}{%
                           \textcolor{cyan}{R#1}%
        }}}}%
    }%
}}

\usepackage{xr-hyper}

\makeatletter
\newcommand*{\addFileDependency}[1]{
  \typeout{(#1)}
  \@addtofilelist{#1}
  \IfFileExists{#1}{}{\typeout{No file #1.}}
}

\makeatother
\newcommand*{\myexternaldocument}[1]{
    \externaldocument{#1}
    \addFileDependency{#1.tex}
    \addFileDependency{#1.aux}
}

\definecolor{cvprblue}{rgb}{0.21,0.49,0.74}
\usepackage[pagebackref,breaklinks,colorlinks,citecolor=cvprblue]{hyperref}
\usepackage[capitalize]{cleveref}
\crefname{section}{Sec.}{Secs.}
\crefname{table}{Table}{Tables}
\crefname{figure}{Fig.}{Figs.}

\ifarxiv \crefname{appendix}{App.}{Apps.}
\else \crefname{appendix}{Suppl.}{Suppls.} \fi

\unless\ifarxiv \myexternaldocument{_supplementary} \fi

\begin{document}
\title{\paperTitle}
\author{\authorBlock}
\maketitle

\begin{abstract}
The goal of selective prediction is to allow an a model to abstain when it may not be able to deliver a reliable prediction, which is important in safety-critical contexts.
Existing approaches to selective prediction typically require access to the internals of a model, require retraining a model or study only unimodal models.
However, the most powerful models (e.g. GPT-4) are typically only available as black boxes with inaccessible internals, are not retrainable by end-users, and are frequently used for multimodal tasks.
We study the possibility of selective prediction for vision-language models in a realistic, black-box setting.
We propose using the principle of \textit{neighborhood consistency} to identify unreliable responses from a black-box vision-language model in question answering tasks.
We hypothesize that given only a visual question and model response, the consistency of the model's responses over the neighborhood of a visual question will indicate reliability.
It is impossible to directly sample neighbors in feature space in a black-box setting.
Instead, we show that it is possible to use a smaller proxy model to approximately sample from the neighborhood.
We find that neighborhood consistency can be used to identify model responses to visual questions that are likely unreliable, even in adversarial settings or settings that are out-of-distribution to the proxy model.
\end{abstract}
\section{Introduction}
\begin{figure}
    \centering
    \includegraphics{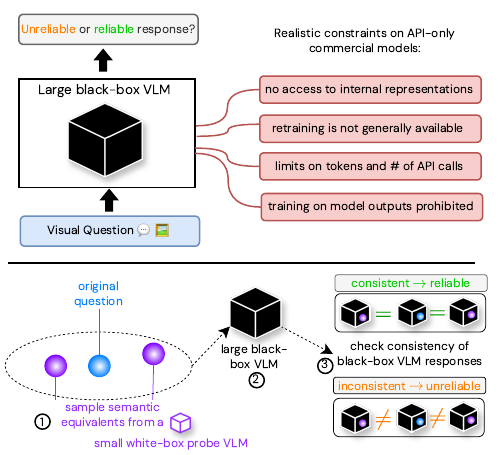}
    \caption{Identifying unreliable responses from an API-only black-box vision-language model (VLM) can be challenging because confidence scores are not always trustworthy, and more sophisticated methods for selective prediction require a level of access to the model that is unavailable.
    We explore the idea of model consistency to identify unreliable model responses in this realistic scenario: a reliable response is one that is consistent across questions that are semantically equivalent but different on the surface.}
    \label{fig:teaser}
\end{figure}
Powerful commercial frontier models are sometimes only available as black boxes accessible through an API \cite{Sun2022BlackBoxTF, 9131769}.
When using these models in high-risk scenarios, it is preferable that the model defers to an expert or abstains from answering rather than deliver an incorrect answer \cite{Geifman2017SelectiveCF}. 
Many approaches for selective prediction \cite{Ziyin2019DeepGL,Geifman2017SelectiveCF} or improving the predictive uncertainty of a model exist, such as ensembling \cite{Lakshminarayanan2016SimpleAS}, gradient-guided sampling in feature space \cite{AttributionBasedConfidenceMetricSwami2019}, retraining the model \cite{van2020uncertainty}, or training a auxiliary module using model predictions \cite{Mozannar2020ConsistentEF}.
Selective prediction has typically been studied in unimodal settings and/or for tasks with a closed-world assumption, such as image classification, and has only recently been studied for multimodal, open-ended tasks such as visual question answering \cite{Whitehead2022ReliableVQ, learning_from_your_peers} (VQA).

In existing deployments, training data is private, model features and gradients are unavailable, retraining is not possible, the number of predictions may be limited by the API, training on model outputs is often prohibited, and queries are open-ended.
In a black-box setting with realistic constraints, how do we identify unreliable predictions from a vision-language model?
 
An intuitive approach is to consider self-consistency: if a human subject is given two semantically equivalent questions, we expect the human subject's answers to the questions to be identical.
A more formal notion of consistency is that given a classifier $f(\cdot)$ and an point $\mathbf{x} \in \mathbb{R}^N$ in feature space, the classifier's predictions over an $\epsilon$-neighborhood of $\mathbf{x}$ should be consistent with $f(\mathbf{x})$ for a small enough $\epsilon$ \cite{AttributionBasedConfidenceMetricSwami2019}.
It is not straightforward to operationalize either of these notions.
How can we scalably obtain ``semantically equivalent'' visual questions to an input visual question? Since we can't access the internal representations of a black-box model, how can we sample from the neighborhood of an input visual question?

First, we study selective prediction on VQA across in-distribution, out-of-distribution, and adversarial inputs using a large VLM.
Next, we describe how rephrasings of a question can be viewed as samples from the $\epsilon$-neighborhood of a visual question pair.
We propose training a visual question generation model as a \textit{probing model} to scalably and cheaply produce rephrasings of visual questions given answers and an image, allowing us to approximately sample from the neighborhood of a visual question pair.
To quantify uncertainty in the answer to a visual question pair, we feed the rephrasings of the question to the black-box VLM, and count the number of rephrasings for which the answer of the VLM remains the same.
Surprisingly, we show that consistency over model-generated ``approximate rephrasings'' is effective at identifying unreliable predictions of a black-box vision-language model, even when the rephrasings are not semantically equivalent and the probing model is an order of magnitude smaller than the black-box model.

Our approach is analogous to consistency over samples taken from the neighborhood of an input sample in feature space, but this method does not require access to the features of the vision-language model.
Furthermore, it does not require a held-out validation set, access to the original training data, or retraining the vision-language model, making it appropriate for black-box uncertainty estimates of a vision-language model.
We conduct a series of experiments testing the effectiveness of consistency over rephrasings for assessing predictive uncertainty using the task of selective visual question answering in a number of settings, including adversarial visual questions, distribution shift, and out of distribution detection.

Our contributions are:
\begin{itemize}
    \item We study the problem of black-box selective prediction for a large vision-language model, using the setting of selective visual question answering.
    \item We show that on in-distribution data, a state-of-the-art large vision-language model is capable of identifying when it does not know the answer to a question, but this ability is severely degraded for out-of-distribution and adversarial visual questions. 
    \item We propose identifying high-risk inputs for visual question answering based on consistency over samples in the neighborhood of a visual question.
    \item We show that consistency defines a different ordering than model confidence / uncertainty over instances in a dataset.
    \item We conduct a series of experiments validating the proposed method on in-distribution, out-of-distribution and adversarial visual questions, and show that our approach even works in the likely setting that the black box model being probed is substantially larger than the probing model.
\end{itemize}

We show that consistency over the rephrasings of a question is correlated with model accuracy on a question and can select slices of a test dataset on which a model can achieve lower risk, reject out of distribution samples, and works well to separate right from wrong answers, even on adversarial and out of distribution inputs.
\textit{Surprisingly, this technique works even though many rephrasings are not literally valid rephrasings of a question.}
Our proposed method is a step towards reliable usage of vision-language models as an API.
\textbf{Limitations:} Due to resource constraints, we study models that might now be considered relatively small ($\leq$ 13B parameters), and our VQG model is ``small'' ($<$ 700M). 

\section{Motivating Experiment}
\begin{figure*}
    \centering
    \includegraphics[width=\textwidth]{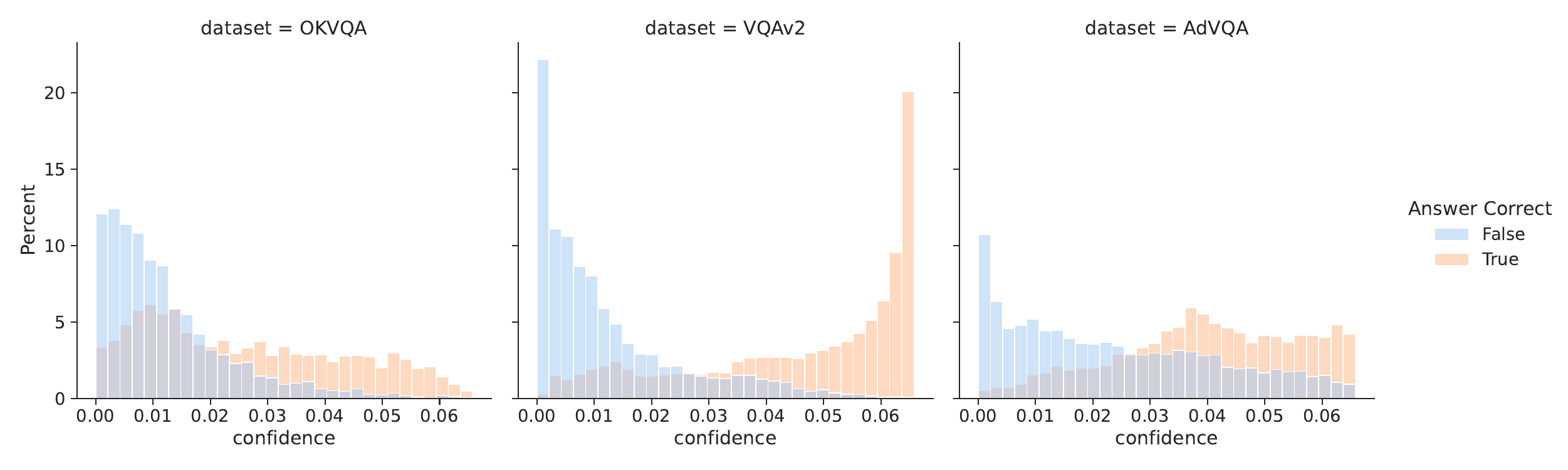}
    \caption{For out of distribution (OKVQA) and adversarial visual (AdVQA) questions, confidence scores alone do not work well to separate right from wrong answers --- many correct answers are low confidence for OOD data, and many wrong answers are high confidence for adversarial data. \textbf{Note: Displayed confidence scores are raw. See Appendix for discussion on calibration.}}
    \label{fig:blip_sep_answers_by_conf}
\end{figure*}
\begin{figure}
    \centering
    \includegraphics[height=1.5in]{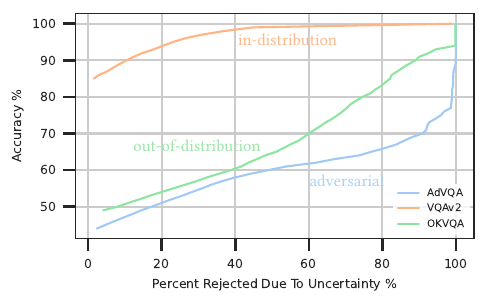}
    \caption{Selective VQA performance of a VLM (BLIP) on three datasets: adversarial (AdVQA), out-of-distribution (OKVQA), and in-distribution (VQAv2). On OOD and adversarial questions, the model has a harder time identifying which questions it should abstain from.}
    \label{fig:blip_risk_cov}
\end{figure}
We empirically examine the predictive uncertainty of a large VLM through the lens of selective visual question answering.
In contrast to the classical VQA setting where a model is forced to answer, a model is allowed to \textit{abstain} from answering in selective VQA.
For safety and reliability, it is important to examine both out-of-distribution and adversarial inputs, on which we expect that the VLM will have a high error rate if forced to answer every out-of-distribution or adversarial question posed to the model. 
However, because the VLM is allowed to abstain, in principle the model can achieve low risk (low error rate) on a \textit{slice} of the dataset corresponding to questions that it knows the answer to. 
In a black-box setting, only the raw confidence scores for the answer candidates are likely to be available, so we use the confidence of the most likely answer as the uncertainty.

In \cref{fig:blip_risk_cov}, we plot the selective visual question answering performance of BLIP \cite{Li2022BLIPBL} finetuned on VQAv2 using the confidence scores on the validation sets of adversarial (AdVQA, \citet{sheng2021human}), out-of-distribution (OKVQA, \citet{Marino2019OKVQAAV}) and in-distribution (VQAv2) datasets.
For the in-distribution dataset (VQAv2), the model is quickly able to identify which questions it is likely to know the answer to, achieving nearly perfect accuracy by rejecting the most uncertain $40\%$ of the dataset.
However, for out-of-distribution and adversarial datasets, the model has a harder time -- after rejecting $50\%$ of the questions, the model still has an error rate of $\approx 40\%$.
The reason for this is evident in \cref{fig:blip_sep_answers_by_conf}, where we plot the distribution of confidence scores for incorrect and correct answers for OOD, in-distribution, and adversarial visual questions.
For in-distribution visual questions, the confidence distribution is bimodal, and incorrect and correct answers are clearly separated by confidence. 
For OOD visual questions, many correctly answered questions are low confidence and difficult to distinguish from incorrectly answered questions.
A similar situation occurs for adversarial visual questions, in which many questions are incorrectly answered with high confidence.

Although the strategy of using model confidence \textit{alone} to detect questions the model cannot answer is effective for in-distribution visual questions, this strategy fails on out-of-distribution and adversarial visual questions.

\section{Method}
\subsection{Task Definition and Background}
\label{sec:task-definition}
Given an image $v$ and question $q$, the task of selective visual question answering is to decide whether a model $f_{VQA}(v,q)$ should predict an answer $a$, or abstain from making a prediction.
A typical solution to this problem is to train a selection function $g(\cdot)$ that produces an abstention score $p_{\mathrm{rej}} \in [0,1]$.
The simplest selection function would be to take the rejection probability $p_{\mathrm{rej}} = 1 - p(a|q,v)$ where $p(a|q,v)$ is the model confidence that $a$ is the answer, and then use a threshold $\tau$ so that the model abstains when $p_{\mathrm{rej}} > \tau $ and predicts otherwise.
A more complex approach taken by \citet{Whitehead2022ReliableVQ} is to train a parametric selection function $g(\mathbf{z}_v, \mathbf{z}_q;\theta)$ where $\mathbf{z}_{v}$ and $\mathbf{z}_q$ are the model's dense representations of the question and image respectively.
The parameters $\theta$ are optimized on a held-out validation set, effectively training a classifier to predict when $f_{VQA}$ will predict incorrectly on an input visual question $v,q$.

In the black box setting, access to the dense representations $\mathbf{z}_v, \mathbf{z}_q$ of the image $v$ and question $q$ is typically forbidden.
Furthermore, even if access to the representation is allowed, a large number of evaluations of $f_{VQA}$ would be needed to obtain the training data for the selection function.
\textbf{Existing methods for selective prediction typically assume and evaluate a fixed set of classes, but for VQA, the label space can shift for each task (differing sets of acceptable answers for different types of questions) or be open-set.}
\begin{enumerate}
    \item The approach should not require access to the black-box model's internal representations of $v,q$.
    \item The approach should be model agnostic, as the architecture of the black-box model is unknown.
    \item The approach should not require a large number of predictions from black-box model to train a selection function, because each usage of the black-box model incurs a financial cost, which can be substantial if large number of predictions are needed to train an auxiliary model.
    \item Similarly, the approach should not require a held-out validation set for calibrating predictions, because this potentially requires a large number of evaluations of the black-box model.
\end{enumerate}
\subsection{Deep Structure and Surface Forms}
Within the field of linguistics, a popular view first espoused by \citet{Chomsky1975-CHOTLS} is that every natural language sentence has both a surface form and a deep structure.
Multiple surface forms can be instances of the same deep structure.
Simply put, multiple sentences that have different words arranged in different orders can mean the same thing.
A rephrasing of a question corresponds to an alternate surface form, but the same deep structure.
We thus expect that the answer to a rephrasing of a question should be the same as the original question.
If the answer to a rephrasing is inconsistent with the answer to an original question, it indicates the model is sensitive to variations in the surface form of the original question.
This indicates the model's understanding of the question is highly dependent on superficial characteristics, making it a good candidate for abstention --- we hypothesize inconsistency on the rephrasings can be used to better quantify predictive uncertainty and reject questions a model has not understood.
\subsection{Rephrasing Generation as Neighborhood Sampling}
\label{sec:rephrasings_as_neighbors}
The idea behind many methods for representation learning is that a good representation should map multiple surface forms close together in feature space.
For example, in contrastive learning, variations in surface form are generated by applying augmentations to an input, and the distance between multiple surface forms is minimized. 
In general, a characteristic of deep representation is that surface forms of an input should be mapped close together in feature space.
Previous work, such as Attribution-Based Confidence \cite{AttributionBasedConfidenceMetricSwami2019} and Implicit Semantic Data Augmentation \cite{Wang2019ImplicitSD}, exploit this by perturbing input samples in \textit{feature space} to explore the neighborhood of an input.
In a black-box setting, we don't have access to the features of the model, so there is no direct way to explore the neighborhood of an input in feature space.
An alternate surface form of the input should be mapped close to the original input in feature space.
Thus, a surface form variation of an input \textit{should} be a neighbor of the input in feature space.
\textit{Generating} a surface form variation of a natural language sentence corresponds to \textit{a rephrasing} of the natural language sentence.
Since a rephrasing of a question is a surface form variation of a question, and surface form variations of an input should be mapped close to the original input in feature space, a rephrasing of a question is analogous to a sample from the neighborhood of a question.
\textbf{We discuss this further in the appendix.}
\subsection{Cyclic Generation of Rephrasings}

\begin{figure*}
    \centering
    \includegraphics[width=\linewidth]{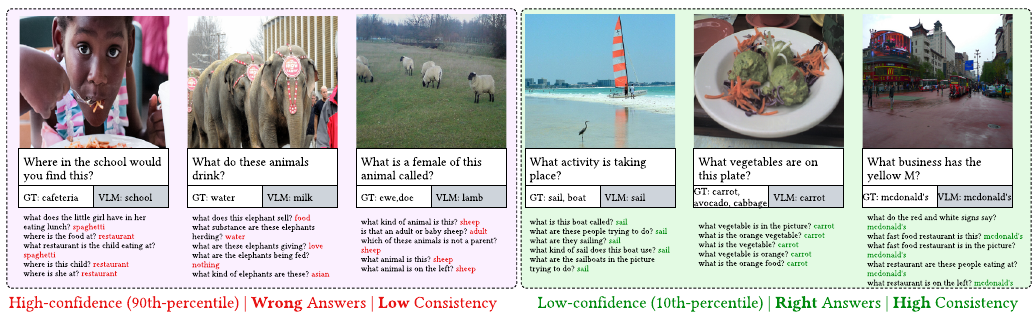}
    \caption{Examples showing the use of model-generated rephrasings to identify errors in model predictions with BLIP as the black box model $f_{BB}$. In the left panel, we show high-confidence answers that wrong, and identified by their low consistency across rephrasings. In the right panel, we show low-confidence answers that are actually correct, identified by their high-confidence across rephrasings.}
    \label{fig:qualitative_samples}
\end{figure*}
A straightforward way to generate a rephrasing of a question is to invert the visual question answering problem, as is done in visual question generation.
Let $p(V), p(Q), p(A)$ be the distribution of images, questions, and answers respectively. 
Visual question generation can be framed as approximating $p(Q|A,V)$, in contrast to visual question answering, which approximates $p(A|Q,V)$. 
We want to probe the predictive uncertainty of a black box visual question answering model $f_{BB}(\cdot)$ on an input visual question pair $v,q$ where $v \sim p(V)$ is an image and $q \sim p(Q)$ is a question.. 
The VQA model $f_{BB}$ approximates $p(A|Q,V)$. 
Let the answer $a$ assigned the highest probability by the VQA model $f_{BB}(\cdot)$ be taken as the prospective answer.
A VQG model $f_{VQG} \approx p(Q|A,V)$ can then be used to generate a rephrasing of an input question $q$.
To see how, consider feeding the highest probability answer $a$ from $f_{BB}(\cdot) \approx p(A|Q,V)$ into $f_{VQG}(\cdot) \approx p(Q|A,V)$ and then sampling a sentence $q^\prime \sim f_{VQG} \approx p(Q|A,V)$ from the visual question generation model.
In the case of an ideal $f_{VQG}(\cdot)$ and perfectly consistent $f_{BB}(\cdot)$, $q^\prime$ should be a generated question for which $p(a|q^\prime, v) \geq p(a_i|q^\prime, v) \forall a_i \in A$, with equality occurring in the case that $a_i = a$. 
So, $q^\prime$ is a question having the same answer as $q$, which is practically speaking, a rephrasing.
We provide an algorithm listing in \cref{consistency-algorithm}.

To summarize, we ask the black box model for an answer to a visual question, then give the predicted answer to a visual question generation model to produce a question $q^\prime$  conditioned on the image $v$ and the answer $a$ by the black box model, which corresponds to a question the VQG model thinks \textit{should} lead to the predicted answer $a$. 
We assume the rephrasings generated by $f_{VQG}$ are good enough, $f_{BB}$ \textit{should} be consistent on the rephrasings, and inconsistency indicates a problem with $f_{BB}$.
In practice, each $q^\prime$ is not guaranteed to be a rephrasing (see \cref{fig:qualitative_samples}) due to the probabilistic nature of the sampling process and because the VQG model is not perfect.
The VQG model can be trained by following any procedure that results in a model approximating $p(a|q,v)$ that is an autoregressive model capable of text generation conditional on multimodal image-text input. 
The training procedure of the VQG model is an implementation detail we discuss in \cref{sec:implementation-details}.

\subsection{Implementation Details}
\label{sec:implementation-details}
We initialize the VQG model $f_{VQG}$ from a BLIP checkpoint pretrained on $129$m image-text pairs, and train it to maximize $p(a|q,v)$ using a standard language modeling loss.
Specifically, we use 
\begin{equation}
\mathcal{L}_{\mathrm{VQG}}=-\sum_{n=1}^N \log P_\theta\left(y_n \mid y_{<n}, a, v\right)
\end{equation}
where $y_1, Y_2, \ldots y_n$ are the tokens of a question $q$ and $a, v$ are the ground-truth answer and image, respectively, from a vqa triplet $(v, q, a)$. We train for 10 epochs, using an AdamW \cite{Loshchilov2017FixingWD} optimizer with a weight decay of 0.05 and decay the learning rate linearly to 0 from 2e-5.
We use a batch size of 64 with an image size of $480\times 480$, and train the model on the VQAv2 training set \cite{balanced_vqa_v2}. 
To sample questions from the VQG model, we use nucleus sampling \cite{Holtzman2019TheCC} with a top-$p$ of 0.9. 

\begin{algorithm}
\SetAlgoLined

\KwIn{$v,q,k$}
\KwData{
$f_{BB}, f_{VQG}$
}
\KwResult{$c \in \mathbb{Q}$: the consistency of $f_{BB}$ over $k$ rephrasings of $v,q$}
$a_0 \longleftarrow f_{BB}(q,v)$ \;
$c \longleftarrow 0$ \;
\For{$i\gets0$ \KwTo $k$}{
    $q^\prime \longleftarrow nucleus\_sample(f_{VQG}(v, a_0)) $\;
    $a^\prime \longleftarrow f_{BB}(q^\prime, v)$ \;
    \If{$a^\prime = a_0$ }{
        $c \longleftarrow c + 1$\;
        }
    }
\KwRet{$c \div k$}
\caption{Probing Predictive Uncertainty of a Black-Box Vision-Language Model}
\label{consistency-algorithm}
\end{algorithm}

\begin{figure*}
    \centering
    \includegraphics[width=\textwidth]{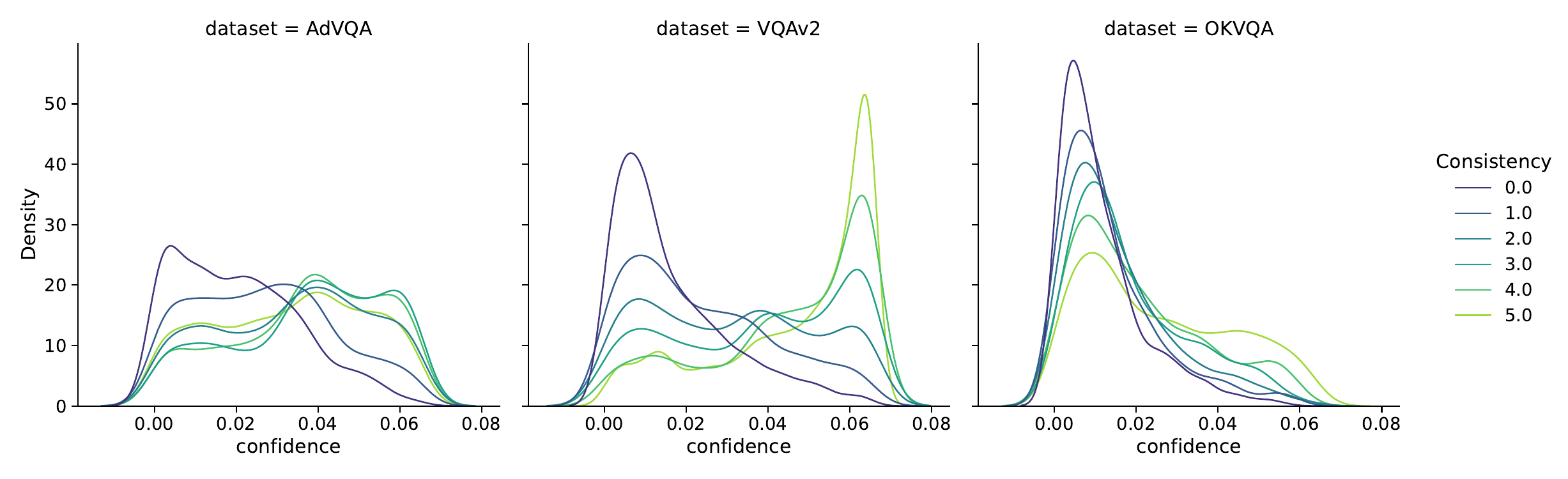}
    \caption{The distribution of confidence scores of $f_{BB}$ at each level of consistency. While higher levels of consistency have a larger proportion of high confidence answers, they also retain a large number of low confidence answers, showing that consistency defines a different ordering over questions than confidence scores alone. BLIP is used as the black-box model $f_{BB}$.}
    \label{fig:confidence-vs-consistency}
\end{figure*}

\section{Experiments}
We conduct a series of experiments probing predictive uncertainty in a black-box visual question answering setting over two large vision-language models and three datasets.
The primary task we use to probe predictive uncertainty is selective visual question answering, which we give a detailed description of in \cref{sec:implementation-details}.
\textbf{\textbf{Futher qualitative examples and results can be found in the appendix.}}

\subsection{Experimental Setup}
\textbf{Black-box Models} The experimental setup requires a black-box VQA model $f_{BB}$ and a rephrasing generator $f_{VQG}$.
We describe the training of the rephrasing generator $f_{VQG}$ in \cref{sec:implementation-details}. 
We choose ALBEF \cite{Li2021AlignBF}, BLIP \cite{Li2022BLIPBL}, and BLIP-2\cite{blip2} as our black-box models.
ALBEF and BLIP have $\approx200\text{m}$ parameters, while the version of BLIP-2 we use is based on the 11B parameter FLAN-T5 \cite{flant5} model.
ALBEF has been pretrained on 14m image-text pairs, while BLIP has been pretrained on over 100m image-text pairs, and BLIP-2 is aligned on 4M images.
We use the official checkpoints provided by the authors, finetuned on Visual Genome \cite{Krishna2016VisualGC} and VQAv2 \cite{balanced_vqa_v2} with $1.4\text{m}$ and $440\text{k}$ training triplets respectively.

\textbf{Datasets} We evaluate in three settings: in-distribution, out-of-distribution, and adversarial. 
For the in-distribution setting, we pairs from the VQAv2 validation set following the selection of \cite{meet2019cycle}. 
For the out-of-distribution setting, we use OK-VQA \cite{Marino2019OKVQAAV}, a dataset for question answering on natural images that requires outside knowledge.
OK-VQA is an natural choice for a out-of-distribution selective prediction task, because many of the questions require external knowledge that a VLM may not have acquired, even through large scale pretraining.
On such questions, a model that knows what it doesn't know should abstain due to lack of requisite knowledge.
Finally, we consider adversarial visual questions in the AdVQA \cite{sheng2021human}. 
We use the official validation splits provided by the authors.
The OK-VQA, AdVQA, and VQAv2 validation sets contain $5$k, $10$k, and $40$k questions respectively.

\begin{figure}
    \centering
    \includegraphics[width=\linewidth]{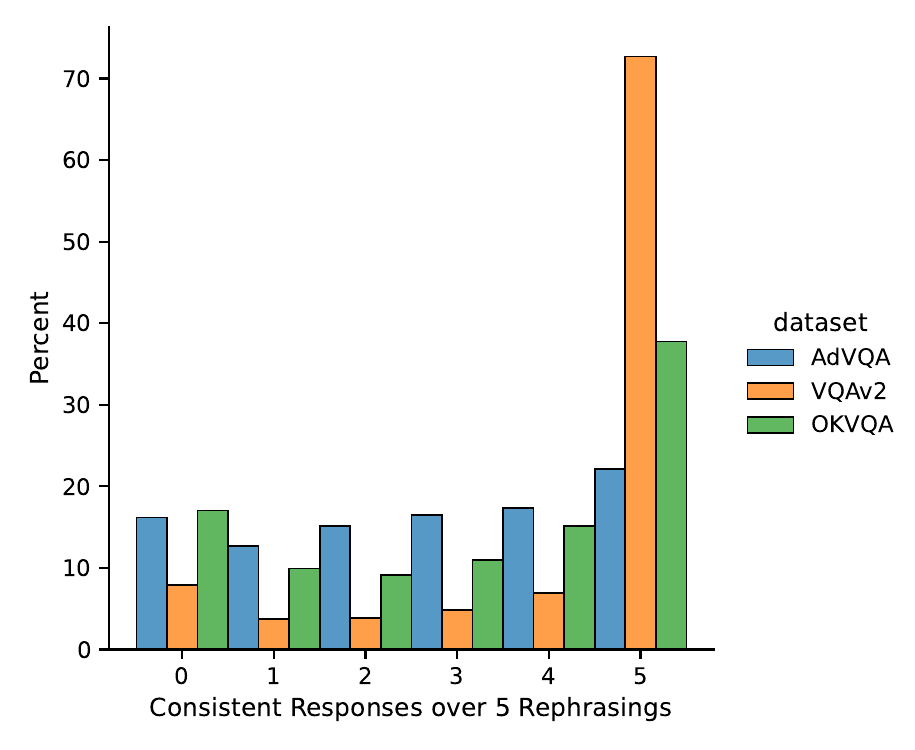}
    \caption{The percentage of each dataset at a given level of consistency. On a well-understood, in-distribution dataset (VQAv2), a large percentage of the questions are at a high consistency level.}
    \label{fig:percent_of_consistency}
\end{figure}

\begin{figure}
    \centering
    \includegraphics[width=\linewidth]{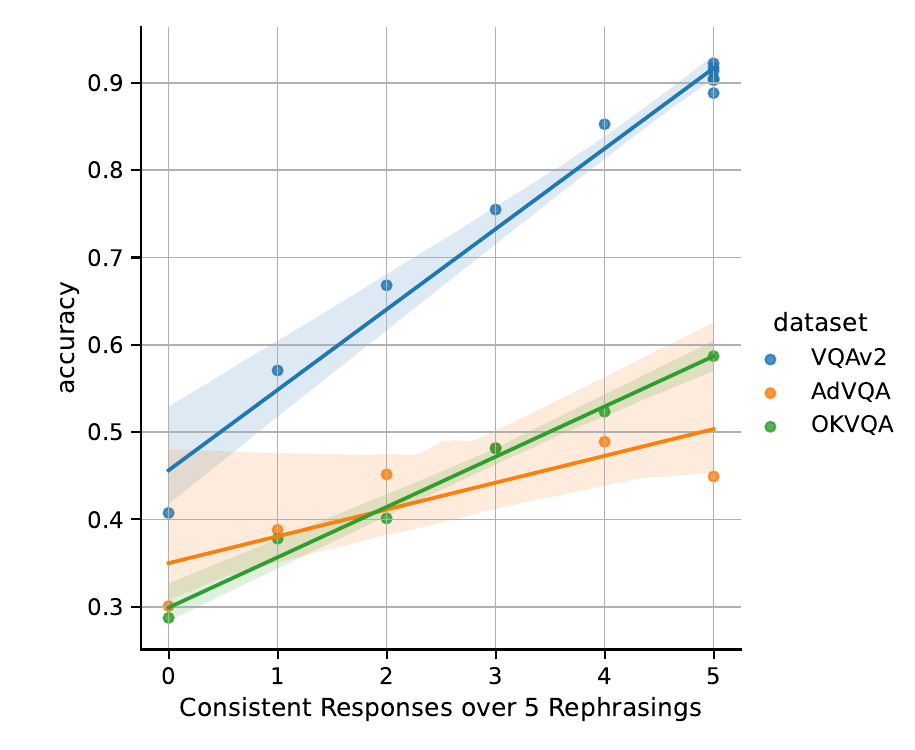}
    \caption{The accuracy of the answers of a VQA model (BLIP) plotted as a function of how consistent each answer was over up to 5 rephrasings of an original question.}
    \label{fig:consistency_vs_accuracy}
\end{figure}
\subsection{Properties of Consistency}

In this section we, analyze properties of the consistency. We are interested in:
\begin{enumerate}
    \item Is increased consistency on rephrasings correlated with model accuracy on the original question?
    \item What does the confidence distribution look like for different levels of consistency?
    \item What is the distribution of consistency across different datasets?
\end{enumerate}
In \cref{fig:consistency_vs_accuracy} we plot the accuracy of the answers when $f_{BB}$ is BLIP by how consistent each answer was over up to 5 rephrasings of an original question. 
We find that consistency over rephrasings is correlated with accuracy across all three datasets, through the correlation is weakest on adversarial data. 
Increased consistency on the rephrasings of a question implies lower risk on the original answer to the original question. 
Next, we examine how the distribution of model confidence varies across consistency levels in \cref{fig:confidence-vs-consistency}.
Across all datasets, slices of a dataset at higher consistency levels also have a greater proportion of high-confidence answers, but \textit{retain a substantial proportion of low confidence answers}. 
This clearly shows that consistency and confidence are not equivalent, and define different orderings on a set of questions and answers.
Put another way, low confidence on a question does not preclude high consistency on a question, and similarly, high confidence on a question does not guarantee the model will be highly consistent on rephrasings of a question.
Finally, plot the percentage of each dataset at a given level of consistency in \cref{fig:percent_of_consistency}.
The in-distribution dataset, VQAv2, has the highest proportion of questions with 5 agreeing neighbors, with all other consistency levels making up the rest of the dataset.
For the out-of-distribution dataset (OKVQA), a substantial proportion of questions ($\approx 40\%$ ) have five agreeing neighbors, with the rest of the dataset shared roughly equally between the other consistency levels.
On the adversarial dataset (AdVQA), the distribution is nearly flat, with equal slices of the dataset at each consistency level. 
One conclusion from this is that higher consistency is not necessarily rarer, and is highly dependent on how well a model understands the data distribution the question is drawn from. 
\subsection{Selective VQA with Neighborhood Consistency}
\begin{figure*}
    \centering
    \includegraphics[width=\textwidth]{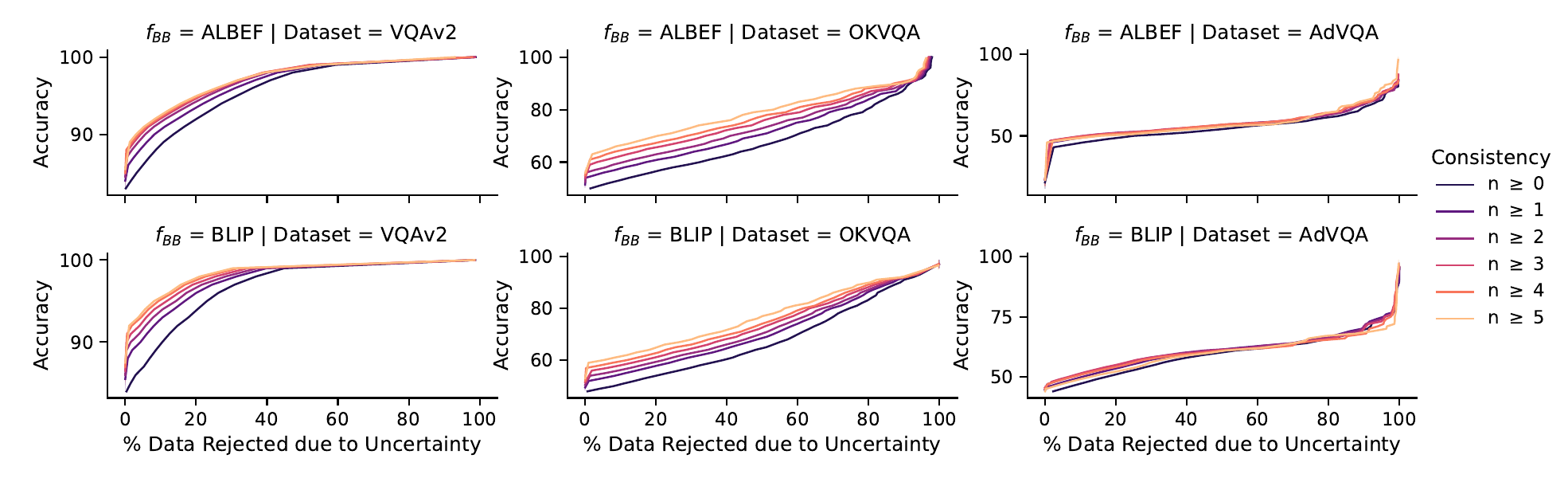}
    \caption{Risk-coverage curves at on slices of test datasets at different levels of consistency. A curve labeled $n \geq k$ shows the risk-coverage tradeoff for a slice of the target dataset where the answers of the model are consistent over at least $k$ rephrasings of an original question. The $n \geq 0$ curve is the baseline. Higher consistency levels identify questions on which a model can achieve lower risk across all datasets.}
    \label{fig:cov_at_risk}
\end{figure*}

\begin{table}
    \begin{minipage}{.48\textwidth}
\adjustbox{max width=\textwidth}{
\begin{tabular}{lllllllllll}
\toprule
$f_{BB}$ & \multicolumn{5}{c}{BLIP} & \multicolumn{5}{c}{ALBEF} \\
\cmidrule(l{0.5em}r{0.5em}){2-6} \cmidrule(l{0.5em}r{0.5em}){7-11}
Risk &           10.0 &           15.0 &           20.0 &           30.0 &           40.0 &           10.0 &           15.0 &           20.0 &          30.0 &          40.0 \\
Consistency &                &                &                &                &                &                &                &                &               &               \\
\midrule
n $\geq$ 0  &           0.11 &           0.18 &           0.25 &            0.4 &           0.61 &           0.08 &           0.14 &           0.21 &          0.41 &          0.68 \\
n $\geq$ 1  &           0.13 &           0.22 &            0.3 &           0.47 &           0.74 &            0.1 &           0.18 &           0.29 &          0.52 &          0.83 \\
n $\geq$ 2  &           0.14 &           0.23 &           0.33 &           0.51 &           0.78 &            0.1 &           0.21 &           0.32 &          0.59 &          0.89 \\
n $\geq$ 3  &           0.16 &           0.26 &           0.37 &           0.56 &           0.84 &           0.12 &           0.23 &           0.37 &          0.66 &          0.97 \\
n $\geq$ 4  &           0.18 &           0.28 &           0.38 &           0.59 &           0.88 &  \textbf{0.13} &           0.26 &           0.42 &          0.71 &  \textbf{1.0} \\
n $\geq$ 5  &  \textbf{0.19} &  \textbf{0.31} &  \textbf{0.44} &  \textbf{0.65} &  \textbf{0.95} &           0.11 &  \textbf{0.33} &  \textbf{0.47} &  \textbf{0.8} &  \textbf{1.0} \\
\bottomrule
\end{tabular}
}
    \vspace{2mm}
    \caption{OK-VQA coverage at a specified risk levels, stratified by consistency levels. $n \geq k$ indicates that the prediction of the model was consistent over at least $k$ rephrasings of the question.}
    \label{tab:okvqa}
    \end{minipage}%
    \hfill
    \begin{minipage}{0.48\textwidth}
\adjustbox{max width=\textwidth}{
\begin{tabular}{lrllllllllr}
\toprule
$f_{BB}$ & \multicolumn{5}{c}{BLIP} & \multicolumn{5}{c}{ALBEF} \\
\cmidrule(l{0.5em}r{0.5em}){2-6} \cmidrule(l{0.5em}r{0.5em}){7-11}
Risk &  20.0 &           30.0 &           40.0 &           50.0 &          56.0 &           20.0 &           30.0 &          40.0 &           50.0 & 60.0 \\
Consistency &       &                &                &                &               &                &                &               &                &      \\
\midrule
n $\geq$ 0  &  0.01 &           0.09 &           0.51 &           0.83 &          0.98 &            0.0 &           0.07 &          0.24 &           0.75 &  1.0 \\
n $\geq$ 1  &  0.01 &  \textbf{0.11} &           0.58 &            0.9 &  \textbf{1.0} &           0.01 &           0.09 &          0.29 &           0.86 &  1.0 \\
n $\geq$ 2  &  0.01 &            0.1 &  \textbf{0.61} &  \textbf{0.93} &  \textbf{1.0} &           0.01 &           0.09 &  \textbf{0.3} &  \textbf{0.89} &  1.0 \\
n $\geq$ 3  &  0.01 &            0.1 &           0.58 &  \textbf{0.93} &  \textbf{1.0} &           0.02 &           0.11 &  \textbf{0.3} &  \textbf{0.89} &  1.0 \\
n $\geq$ 4  &  0.01 &           0.08 &           0.55 &           0.92 &  \textbf{1.0} &           0.02 &           0.11 &  \textbf{0.3} &           0.87 &  1.0 \\
n $\geq$ 5  &  0.01 &           0.04 &           0.53 &           0.87 &  \textbf{1.0} &  \textbf{0.04} &  \textbf{0.12} &          0.27 &           0.84 &  1.0 \\
\bottomrule
\end{tabular}
}
    \vspace{2mm}
    \caption{AdVQA coverage at a specified risk levels, stratified by consistency levels.
    $n \geq k$ indicates that the prediction of the model was consistent over at least $k$ rephrasings of the question.
    }
    \label{tab:advqa}
    \end{minipage}
\end{table}

\begin{figure*}
    \centering
    \includegraphics[width=\textwidth]{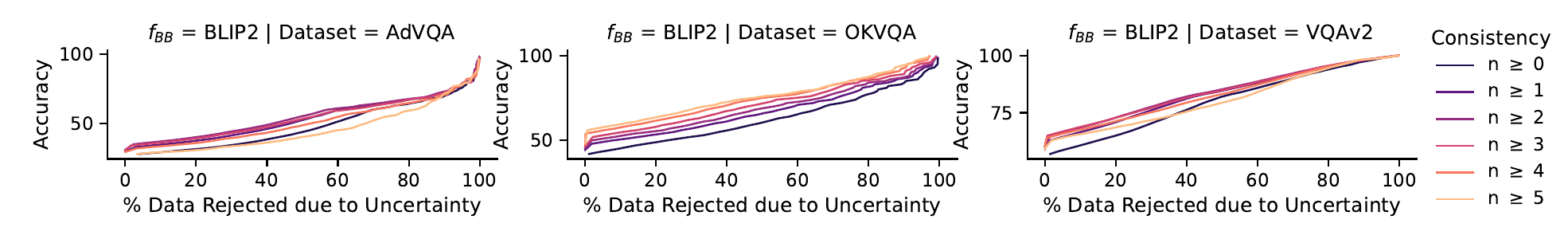}
    \caption{Risk-coverage curves when $f_{VQG}$ (200m parameters) is substantially smaller than $f_{BB}$ (11B). Even in this scenario, $f_{VQG}$ can reliably identify high-risk questions based on consistency.}
    \label{fig:blip2_risk_cov}
\end{figure*}

We turn to the question of whether consistency over rephrasings is useful in the setting of selective visual question answering.
\begin{enumerate}
    \item Can consistency select slices of a test dataset which a model understands well (achieves lower risk), or alternatively, identify questions the model doesn't understand, and should reject (high risk)?
    \item How well does consistency over rephrasings work to identify low / high risk questions in out-of-distribution and adversarial settings?\
    \item What happens when the question generator is much smaller than the black-box model?
\end{enumerate}
To analyze how useful consistency is for separating low-risk from high-risk inputs, we use the task of selective visual question answering. 
In \cref{fig:cov_at_risk} we plot risk-coverage curves for in-distribution, out-of-distribution, and adversarial visual questions.
Each curve shows the risk-coverage tradeoff for questions at a level of consistency.
For example, a curve labeled as $n \geq 3$ shows the risk-coverage tradeoff for questions on which 3 or more neighbors (rephrasings) were consistent with the original answer. 
Hence, the $n \geq 0$ curve is a baseline representing the risk-coverage curve for any question, regardless of consistency.
If greater consistency over rephrasings is indicative over lower risk (and a higher probability the model knows the answer), we expect to see that the model should be able to achieve lower risk on slices of a dataset that the model is more consistent on. 
On in-distribution visual questions (VQAv2), the model achieves lower risk at equivalent coverage for slices of the dataset that have higher consistency levels. 
A similar situation holds for the out-of-distribution dataset, OKVQA, and the adversarial dataset AdVQA. 
In general, the model is able to achieve lower risk on slices of a dataset on which the consistency of the model over rephrasings is higher. 
In \cref{tab:advqa} and \cref{tab:okvqa}, we show risk-coverage information in tabular form for AdVQA and OK-VQA at specific risk levels.
Finally, in \cref{fig:blip_risk_cov}, we show that our approach works even when there is a large size difference between the black-box model and the question generator.
\section{Related Work}
\subsection{Selective Prediction}
Deep models with a reject option have been studied in the context of unimodal classification and regression \cite{Geifman2017SelectiveCF,Geifman2019SelectiveNetAD,Ziyin2019DeepGL} for some time, and more recently for the open-ended task of question answering \cite{Kamath2020SelectiveQA}.
Deep models with a reject option in the context of visual question answering were first explored by \citet{Whitehead2022ReliableVQ}.
They take the approach of training a selection function using features from the model and a held-out validation set to make the decision of whether to predict or abstain.
\citet{learning_from_your_peers} takes an alternate approach by training models on different dataset slices.
The problem of eliciting truthful information from a language model \cite{Lin2021TruthfulQAMH} is closely related to selective prediction for VQA.
In both settings, the model must avoid providing false information in response to a question.
\subsection{Self-Consistency}
\citet{AttributionBasedConfidenceMetricSwami2019} introduced the idea of using consistency over the predictions of a model to quantify the predictive uncertainty of the model. 
Their Attribution Based Confidence (ABC) metric is based on using guidance from feature attributions, specifically Integrated Gradients \cite{Sundararajan2017AxiomaticAF} to perturb samples in feature space, then using consistency over the perturbed samples to quantify predictive uncertainty.
\citet{meet2019cycle} show that VQA models are not robust to linguistic variations in a sentence by demonstrating inconsistency of the answers of multiple VQA models over human-generated rephrasings of a sentence.
Similarly, \citet{Selvaraju2020SQuINTingAV} show that the answers of VQA models to more complex reasoning questions are inconsistent with the answers to simpler perceptual questions whose answers should entail the answer to the reasoning question. 
We connect these ideas to hypothesize that inconsistency on linguistic variations of a visual question is indicative of more superifical understanding of the content of the question, and therefore a higher chance of being wrong when answering the question.
\subsection{Robustness of VQA Models}
VQA models have been shown to lack robustness, and severely prone to overfitting on dataset-specific correlations rather than learning to answer questions. 
The VQA-CP \cite{Agrawal2017DontJA} task showed that VQA models often use linguistic priors to answer questions (e.g. the sky is usually blue), rather than looking at the image.
\citet{Dancette2021BeyondQB} showed that VQA models often use simple rules based on co-occurences of objects with noun phrases to answer questions. 
The existence of adversarial visual questions has also been demonstrated by \cite{sheng2021human}, who used an iterative model-in-the-loop process to allow human annotators to attack state-of-the-art 
While VQA models are approaching human-level performance on the VQAv2 benchmark \cite{balanced_vqa_v2}, their performance on more complex VQA tasks such as OK-VQA \cite{Marino2019OKVQAAV} lags far behind human performance.
\section{Conclusion}
The capital investment required to train large, powerful models on massive amounts of data means that there is a strong commercial incentive to keep the weights and features of a model private while making the model accessible through an API. 
Using these models in low-risk situations is not problematic, but using black-box models in situations where mistakes can have serious consequences is dangerous. 
At the same time, the power of these black-box models makes using them very appealing.

In this paper, we explore a way to judge the reliability of the answer of a black-box visual question answering model by assessing the consistency of the model's answer over rephrasings of the original question, which we generate dynamically using a VQG model. 
We show that this is analogous to the technique of consistency over neighborhood samples, which has been used in white-box settings for self-training as well as predictive uncertainty.
We conduct experiments on in-distribution, out-of-distribution, and adversarial settings, and show that consistency over rephrasings is correlated with model accuracy, and predictions of a model that are highly consistent over rephrasings are more likely to be correct. 
Hence, consistency over rephrasings constitutes an effective first step for using a black-box visual question answering model reliably by identifying queries that a black-box model may not know the answer to.

{\small
\bibliographystyle{ieeenat_fullname}
\bibliography{11_references}
}

\ifarxiv \clearpage \appendix \onecolumn
\section{Detailed Risk-Coverage Data}
In \cref{tab:advqa_complete,tab:okvqa_complete,tab:vqav2_albef,tab:vqav2_blip}, we show more granular risk-coverage curves across all three evaluated datasets and both black-box models.
\begin{table}[]
    \centering
    \adjustbox{max width=\linewidth}{
\begin{tabular}{lrrllllllllllllllllll}
\toprule
$f_{BB}$ & \multicolumn{10}{c}{BLIP} & \multicolumn{10}{c}{ALBEF} \\
Risk & 0.0  & 5.0  &           10.0 &           15.0 &           20.0 &           25.0 &           30.0 &          35.0 &           40.0 &          45.0 &           0.0  &           5.0  &           10.0 &           15.0 &           20.0 &           25.0 &          30.0 &           35.0 &          40.0 &          45.0 \\
Consistency &      &      &                &                &                &                &                &               &                &               &                &                &                &                &                &                &               &                &               &               \\
\midrule
n $\geq$ 0  &  0.0 &  0.0 &           0.11 &           0.18 &           0.25 &           0.32 &            0.4 &          0.49 &           0.61 &          0.77 &           0.02 &           0.03 &           0.08 &           0.14 &           0.21 &            0.3 &          0.41 &           0.53 &          0.68 &          0.85 \\
n $\geq$ 1  &  0.0 &  0.0 &           0.13 &           0.22 &            0.3 &           0.38 &           0.47 &          0.59 &           0.74 &          0.89 &           0.02 &           0.04 &            0.1 &           0.18 &           0.29 &            0.4 &          0.52 &           0.66 &          0.83 &          0.97 \\
n $\geq$ 2  &  0.0 &  0.0 &           0.14 &           0.23 &           0.33 &           0.42 &           0.51 &          0.63 &           0.78 &          0.94 &           0.03 &           0.04 &            0.1 &           0.21 &           0.32 &           0.45 &          0.59 &           0.73 &          0.89 &  \textbf{1.0} \\
n $\geq$ 3  &  0.0 &  0.0 &           0.16 &           0.26 &           0.37 &           0.45 &           0.56 &          0.68 &           0.84 &  \textbf{1.0} &           0.03 &           0.05 &           0.12 &           0.23 &           0.37 &           0.51 &          0.66 &           0.83 &          0.97 &  \textbf{1.0} \\
n $\geq$ 4  &  0.0 &  0.0 &           0.18 &           0.28 &           0.38 &           0.48 &           0.59 &          0.74 &           0.88 &  \textbf{1.0} &  \textbf{0.04} &  \textbf{0.06} &  \textbf{0.13} &           0.26 &           0.42 &           0.55 &          0.71 &           0.88 &  \textbf{1.0} &  \textbf{1.0} \\
n $\geq$ 5  &  0.0 &  0.0 &  \textbf{0.19} &  \textbf{0.31} &  \textbf{0.44} &  \textbf{0.54} &  \textbf{0.65} &  \textbf{0.8} &  \textbf{0.95} &  \textbf{1.0} &  \textbf{0.04} &  \textbf{0.06} &           0.11 &  \textbf{0.33} &  \textbf{0.47} &  \textbf{0.63} &  \textbf{0.8} &  \textbf{0.93} &  \textbf{1.0} &  \textbf{1.0} \\
\bottomrule
\end{tabular}
}
    \caption{More granular risk-coverage data for OK-VQA.}
    \label{tab:okvqa_complete}
\end{table}

\begin{table}[]
    \centering
    \adjustbox{max width=\linewidth}{
\begin{tabular}{lrllllllllllllllllr}
\toprule
$f_{BB}$ & \multicolumn{9}{c}{BLIP} & \multicolumn{9}{c}{ALBEF} \\
Risk &  20.0 &           25.0 &           30.0 &           35.0 &           40.0 &          45.0 &           50.0 &          55.0 &          56.0 &           20.0 &           25.0 &           30.0 &           35.0 &          40.0 &          45.0 &           50.0 &          55.0 & 60.0 \\
Consistency &       &                &                &                &                &               &                &               &               &                &                &                &                &               &               &                &               &      \\
\midrule
n $\geq$ 0  &  0.01 &  \textbf{0.04} &           0.09 &           0.23 &           0.51 &          0.69 &           0.83 &          0.95 &          0.98 &            0.0 &           0.04 &           0.07 &           0.12 &          0.24 &          0.46 &           0.75 &          0.92 &  1.0 \\
n $\geq$ 1  &  0.01 &  \textbf{0.04} &  \textbf{0.11} &  \textbf{0.27} &           0.58 &          0.76 &            0.9 &  \textbf{1.0} &  \textbf{1.0} &           0.01 &           0.05 &           0.09 &           0.15 &          0.29 &          0.55 &           0.86 &  \textbf{1.0} &  1.0 \\
n $\geq$ 2  &  0.01 &  \textbf{0.04} &            0.1 &           0.25 &  \textbf{0.61} &          0.79 &  \textbf{0.93} &  \textbf{1.0} &  \textbf{1.0} &           0.01 &           0.05 &           0.09 &           0.15 &  \textbf{0.3} &          0.59 &  \textbf{0.89} &  \textbf{1.0} &  1.0 \\
n $\geq$ 3  &  0.01 &  \textbf{0.04} &            0.1 &           0.25 &           0.58 &  \textbf{0.8} &  \textbf{0.93} &  \textbf{1.0} &  \textbf{1.0} &           0.02 &           0.06 &           0.11 &           0.17 &  \textbf{0.3} &  \textbf{0.6} &  \textbf{0.89} &  \textbf{1.0} &  1.0 \\
n $\geq$ 4  &  0.01 &           0.02 &           0.08 &           0.24 &           0.55 &          0.77 &           0.92 &  \textbf{1.0} &  \textbf{1.0} &           0.02 &           0.06 &           0.11 &           0.16 &  \textbf{0.3} &  \textbf{0.6} &           0.87 &  \textbf{1.0} &  1.0 \\
n $\geq$ 5  &  0.01 &           0.01 &           0.04 &  \textbf{0.27} &           0.53 &          0.72 &           0.87 &  \textbf{1.0} &  \textbf{1.0} &  \textbf{0.04} &  \textbf{0.07} &  \textbf{0.12} &  \textbf{0.18} &          0.27 &          0.53 &           0.84 &  \textbf{1.0} &  1.0 \\
\bottomrule
\end{tabular}}
    \caption{More granular risk-coverage data for AdVQA.}
    \label{tab:advqa_complete}
\end{table}

\begin{table}[]
    \centering
    \adjustbox{max width=\linewidth}{
    \begin{tabular}{lrllllllllllllll}
\toprule
$f_{BB}$ & \multicolumn{15}{l}{BLIP} \\
risk &  0.0  &          1.0  &           2.0  &           3.0  &           4.0  &           5.0  &           6.0  &           7.0  &           8.0  &          9.0  &          10.0 &          11.0 &          12.0 &          13.0 &          14.0 \\
Consistency &       &               &                &                &                &                &                &                &                &               &               &               &               &               &               \\
\midrule
n $\geq$ 0  &  0.01 &          0.55 &           0.63 &           0.69 &           0.74 &           0.77 &            0.8 &           0.82 &           0.85 &          0.88 &           0.9 &          0.91 &          0.93 &          0.95 &          0.97 \\
n $\geq$ 1  &  0.01 &           0.6 &           0.69 &           0.76 &            0.8 &           0.83 &           0.86 &            0.9 &           0.92 &          0.94 &          0.96 &          0.98 &          0.99 &  \textbf{1.0} &  \textbf{1.0} \\
n $\geq$ 2  &  0.01 &          0.63 &           0.72 &           0.78 &           0.83 &           0.86 &           0.89 &           0.92 &           0.94 &          0.96 &          0.98 &  \textbf{1.0} &  \textbf{1.0} &  \textbf{1.0} &  \textbf{1.0} \\
n $\geq$ 3  &  0.01 &          0.66 &           0.75 &           0.81 &           0.85 &           0.88 &           0.92 &           0.94 &           0.96 &          0.98 &  \textbf{1.0} &  \textbf{1.0} &  \textbf{1.0} &  \textbf{1.0} &  \textbf{1.0} \\
n $\geq$ 4  &  0.01 &          0.68 &           0.77 &           0.83 &           0.87 &           0.91 &           0.93 &  \textbf{0.96} &           0.98 &          0.99 &  \textbf{1.0} &  \textbf{1.0} &  \textbf{1.0} &  \textbf{1.0} &  \textbf{1.0} \\
n $\geq$ 5  &  0.01 &  \textbf{0.7} &  \textbf{0.79} &  \textbf{0.84} &  \textbf{0.88} &  \textbf{0.92} &  \textbf{0.94} &  \textbf{0.96} &  \textbf{0.99} &  \textbf{1.0} &  \textbf{1.0} &  \textbf{1.0} &  \textbf{1.0} &  \textbf{1.0} &  \textbf{1.0} \\
\bottomrule
\end{tabular}
}
    \caption{Granular risk-coverage data for VQAv2 with BLIP as $f_{BB}$.}
    \label{tab:vqav2_blip}
\end{table}

\begin{table}[]
    \centering
    \adjustbox{max width=\linewidth}{
    \begin{tabular}{lrllllllllllllll}
\toprule
$f_{BB}$ & \multicolumn{15}{l}{ALBEF} \\
risk &  0.0  &          1.0  &           2.0  &           3.0  &           4.0  &           5.0  &           6.0  &           7.0  &           8.0  &          9.0  &          10.0 &          11.0 &          12.0 &          13.0 &          14.0 \\
Consistency &       &               &                &                &                &                &                &                &                &               &               &               &               &               &               \\
\midrule
n $\geq$ 0  &  0.01 &          0.55 &           0.63 &           0.69 &           0.74 &           0.77 &            0.8 &           0.82 &           0.85 &          0.88 &           0.9 &          0.91 &          0.93 &          0.95 &          0.97 \\
n $\geq$ 1  &  0.01 &           0.6 &           0.69 &           0.76 &            0.8 &           0.83 &           0.86 &            0.9 &           0.92 &          0.94 &          0.96 &          0.98 &          0.99 &  \textbf{1.0} &  \textbf{1.0} \\
n $\geq$ 2  &  0.01 &          0.63 &           0.72 &           0.78 &           0.83 &           0.86 &           0.89 &           0.92 &           0.94 &          0.96 &          0.98 &  \textbf{1.0} &  \textbf{1.0} &  \textbf{1.0} &  \textbf{1.0} \\
n $\geq$ 3  &  0.01 &          0.66 &           0.75 &           0.81 &           0.85 &           0.88 &           0.92 &           0.94 &           0.96 &          0.98 &  \textbf{1.0} &  \textbf{1.0} &  \textbf{1.0} &  \textbf{1.0} &  \textbf{1.0} \\
n $\geq$ 4  &  0.01 &          0.68 &           0.77 &           0.83 &           0.87 &           0.91 &           0.93 &  \textbf{0.96} &           0.98 &          0.99 &  \textbf{1.0} &  \textbf{1.0} &  \textbf{1.0} &  \textbf{1.0} &  \textbf{1.0} \\
n $\geq$ 5  &  0.01 &  \textbf{0.7} &  \textbf{0.79} &  \textbf{0.84} &  \textbf{0.88} &  \textbf{0.92} &  \textbf{0.94} &  \textbf{0.96} &  \textbf{0.99} &  \textbf{1.0} &  \textbf{1.0} &  \textbf{1.0} &  \textbf{1.0} &  \textbf{1.0} &  \textbf{1.0} \\
\bottomrule
\end{tabular}
    }
    \caption{Granular risk-coverage data for VQAv2 with ALBEF as $f_{BB}$.}
    \label{tab:vqav2_albef}
\end{table}
\section{Inference Details}
For both BLIP and ALBEF, we follow the original inference procedures. 
Both models have an encoder-decoder architecture and VQA is treated as a text-to-text task.
We use the rank-classification approach \cite{Brown2020LanguageMA} to allow the autoregressive decoder of the VLM to predict an answer for a visual question.
Concretely, let $\mathcal{A} = \{ a_1, a_2, a_3, \ldots a_k \}$ be a list of length $k$ for a dataset consisting of the most frequent ground-truth answers.
These answer lists are standardized and distributed by the authors of the datasets themselves.
We use the standard answer lists for each dataset.
Next, let $v,q$ be a visual question pair and let $f_{BB}$ be a VQA model.
Recall that $f_{BB}$ is a language model defining a distribution $p(a|q,v)$, and is thus able to assign a score to each $a_i \in \mathcal{A}$.
We take the highest probability $a_k$
\begin{equation}
    \underset{a_k \in \mathcal{A}}{\operatorname{max}}~f_{BB}(v,q,a_k) = \underset{a_k \in \mathcal{A}}{\operatorname{max}} p(a_k | v,q)
\end{equation}
as the predicted answer for a question.
This is effectively asking the model to rank each of the possible answer candidates, turning the open-ended VQA task into a very large multiple choice problem.
Note that the highest probability $a_k \in \mathcal{A}$ is \textit{not} necessarily the answer that would be produced by $f_{BB}\sim p(a|v,q)$ in an unconstrained setting such as stochastic decoding.
However, for consistency with previous work, we use the rank classification approach.

Visual question answering is thus treated differently when using large autoregressive vision-language models compared to non-autoregressive odels. In traditional approaches, VQA is treated as a classification task, and a standard approach used in older, non-autoregressive vision-language models such as ViLBERT \cite{Lu2019ViLBERTPT} is to train a MLP with a cross-entropy loss with each of the possible answers as a class.

\section{Hallucinations}
\begin{figure}
    \centering
    \includegraphics[width=\linewidth]{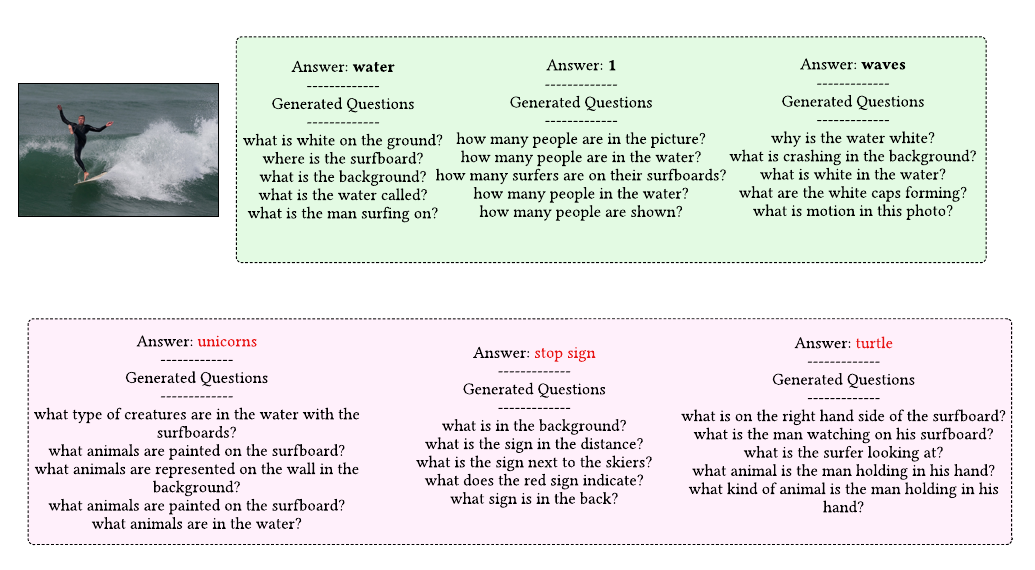}
    \caption{The rephrasing generator $f_{VQG}$ can hallucinate questions that imagine not present in the context of the image.}
    \label{fig:hallucinations}
\end{figure}
We describe a peculiar mode of the rephrasing generator $f_{VQG}$ in this section.
When an answer is out-of-context for a given image, the rephrasing generator $f_{VQG}$ will generate questions premised on the out-of-context answer.
For example, in \cref{fig:hallucinations}, we show that if an out-of-context answer such as ``unicorn'' for the surfing image in \cref{fig:hallucinations} is provided to $f_{VQG}$ for cycle-consistent rephrasing generation, $f_{VQG}$ will generate questions such as  ``what animals are in the water'', assuming that there are unicorns in the water, though this is implausible. 
A more correct question would have been something such as ``what animals are not present?'' A likely reason $f_{VQG}$ cannot handle these cases well is because $f_{VQG}$ is trained on a VQA dataset to approximate $p(q|v,a)$, and traditional VQA datasets have very few counterfactual questions such as these.

This is not specific to the $f_{VQG}$ used in our framework, and should apply to any question generator trained in this manner. 
It does reveal that even large VLMs pretrained on a massive amount of image-text pairs have a superficial understanding of counterfactuals, and possibly other properties of language.

\section{Are the rephrasings really rephrasings?}
\label{sec:rephrasings-discussion}
As visible in \cref{fig:qualitative_samples}, some of the rephrasings are not literally rephrasings of the original question.
It may be more correct to call the rephrasings pseudo-rephrasings, in the same way that generated labels are referred to as pseudolabels in the semi-supervised learning literature \cite{Liu2021CycleSF}.
However, the pseudo-rephrasings seem to be \textit{good enough} that inconsistency over the pseudo-rephrasings indicates potentially unreliable predictions from $f_{BB}$. 

\begin{figure}
    \centering
    \includegraphics{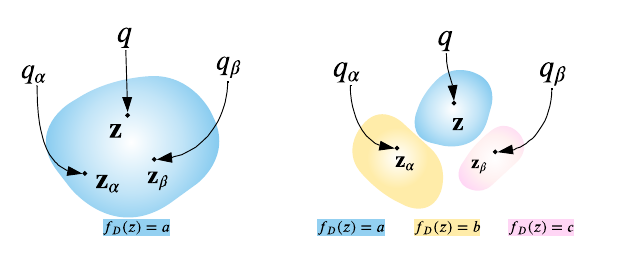}
    \caption{See \cref{sec:rephrasings-discussion} for an explanation of the figure.}
    \label{fig:embedding}
\end{figure}

Why does this work? Decompose $f_{BB}$ as $f_{BB} = f_D(f_E(v,q))$, where $f_E(v,z) = \mathbf{z}$ is the encoder that maps a visual question pair $v,q$ to a dense representation $\mathbf{z}$, and $f_D(\mathbf{z}) = a$ is the decoder that maps the dense representation $\mathbf{z}$ to an answer.
For two rephrasings $q_\alpha, q_\beta$ of a question $q$, the model will be consistent over the rephrasings if all the rephrasings are embedded onto a subset of the embedding space that $f_D$ assigns the same answer $a$.
This is the situation we depict on the left side of \cref{fig:embedding}.

On the other hand, if $q_\alpha$ and $q_\beta$ are embedded into parts of the embedding space that $f_D$ assigns them different answers, the answers will not be consistent (right side of \cref{fig:embedding}).
Thus, whether a $q_\alpha,q_\beta$ are linguistically valid rephrasings does not matter so much as if $q_\alpha, q_\beta$ \textit{should} technically have the same answer as the original question $q$.
Of course, it is true that the answer to a linguistically valid rephrasing should be the same as the same as the answer to the question being rephrased.
However, for any question, there are many other questions that have the same answer but are \textit{not} rephrasings of the original question.

\section{Calibration}
\label{sec:calibration}
The confidence scores in \cref{fig:blip_sep_answers_by_conf,fig:confidence-vs-consistency} are the raw scores from the logits of the VQA model, in this case BLIP.
Recall that the models under consideration are autoregressive models that approximate a probability distribution $p(a|v,q)$, where $a$ can take on an infinite number of values --- the model must be able to assign a score to any natural language sentence. 
The raw distribution of confidence scores is clearly truncated in the sense that all scores appear to lie in the interval $[0, 0.07 ]$.
We apply temperature scaling \cite{10.1145/1102351.1102430} to assess how well the confidence scores are calibrated.
In temperature scaling, the logits of a model are multiplied by a parameter $\tau$.
This is rank-preserving, and yields confidence scores that are more directly interpretable.
In our case, we can use it to rescale the model logits into the interval $[0,1]$ and analyze the \textit{Adaptive Calibration Error} \cite{Nixon2019MeasuringCI} of the model's predictions.
We grid search the $\tau$ that minimizes the Adaptive ECE directly on the model predictions, and show the results in \cref{tab:calibration_okvqa,tab:calibration_vqa,tab:calibration_advqa}.
The Adaptive Calibration Error is lowest on the in-distribution dataset, highest on the adversarial dataset, and second highest on the out-of-distribution dataset.
Notably, the model is systematically overconfident on adversarial samples, but not on out-of-distribution samples.
This suggests that calibration is not the \textit{only} problem in selective prediction.
\begin{table}[]
    \centering
\begin{tabular}{lrrrr}
\toprule
{} &  Raw Confidence &  Accuracy &  Scaled Confidence &  Error \\
percentile &                 &           &                    &        \\
\midrule
0          &           0.020 &     0.477 &              0.390 &  0.087 \\
10         &           0.022 &     0.507 &              0.430 &  0.077 \\
20         &           0.024 &     0.540 &              0.473 &  0.067 \\
30         &           0.026 &     0.573 &              0.522 &  0.051 \\
40         &           0.029 &     0.604 &              0.577 &  0.026 \\
50         &           0.032 &     0.647 &              0.643 &  0.004 \\
60         &           0.036 &     0.699 &              0.723 &  0.024 \\
70         &           0.041 &     0.766 &              0.819 &  0.053 \\
80         &           0.047 &     0.831 &              0.934 &  0.104 \\
90         &           0.054 &     0.909 &              1.000 &  0.091 \\
\bottomrule
\end{tabular}
    \caption{Calibration of BLIP on OK-VQA. For scaling, a temperature of $19.9$ is used.}
    \label{tab:calibration_okvqa}
\end{table}

\begin{table}
    \centering
\begin{tabular}{lrrrr}
\toprule
{} &  Raw Confidence &  Accuracy &  Scaled Confidence &  Error \\
percentile &                 &           &                    &        \\
\midrule
0          &           0.042 &     0.837 &              0.841 &  0.004 \\
10         &           0.047 &     0.898 &              0.926 &  0.028 \\
20         &           0.051 &     0.938 &              1.000 &  0.062 \\
30         &           0.055 &     0.968 &              1.000 &  0.032 \\
40         &           0.058 &     0.984 &              1.000 &  0.016 \\
50         &           0.060 &     0.994 &              1.000 &  0.006 \\
60         &           0.062 &     0.998 &              1.000 &  0.002 \\
70         &           0.064 &     0.999 &              1.000 &  0.001 \\
80         &           0.065 &     1.000 &              1.000 &  0.000 \\
90         &           0.065 &     0.999 &              1.000 &  0.001 \\
\bottomrule
\end{tabular}
    \caption{Calibration of BLIP on VQAv2. For scaling, a temperature of $19.3$ is used.}
    \label{tab:calibration_vqa}
\end{table}

\begin{table}[]
    \centering
\begin{tabular}{lrrrr}
\toprule
{} &  Raw Confidence &  Accuracy &  Scaled Confidence &  Error \\
percentile &                 &           &                    &        \\
\midrule
0          &           0.032 &     0.430 &              0.637 &  0.206 \\
10         &           0.035 &     0.472 &              0.703 &  0.231 \\
20         &           0.039 &     0.510 &              0.769 &  0.259 \\
30         &           0.042 &     0.547 &              0.834 &  0.287 \\
40         &           0.045 &     0.580 &              0.897 &  0.317 \\
50         &           0.048 &     0.601 &              0.956 &  0.355 \\
60         &           0.051 &     0.618 &              1.000 &  0.382 \\
70         &           0.055 &     0.636 &              1.000 &  0.364 \\
80         &           0.058 &     0.655 &              1.000 &  0.345 \\
90         &           0.062 &     0.693 &              1.000 &  0.307 \\
\bottomrule
\end{tabular}
    \caption{Calibration of BLIP on AdVQA. For scaling, a temperature of $12.5$ is used.}
    \label{tab:calibration_advqa}
\end{table}
\section{More Rephrasings Examples}
We show more examples of generated rephrasings by \cref{fig:extended_qualitative}.
\begin{figure}
    \centering
    \includegraphics[width=\linewidth]{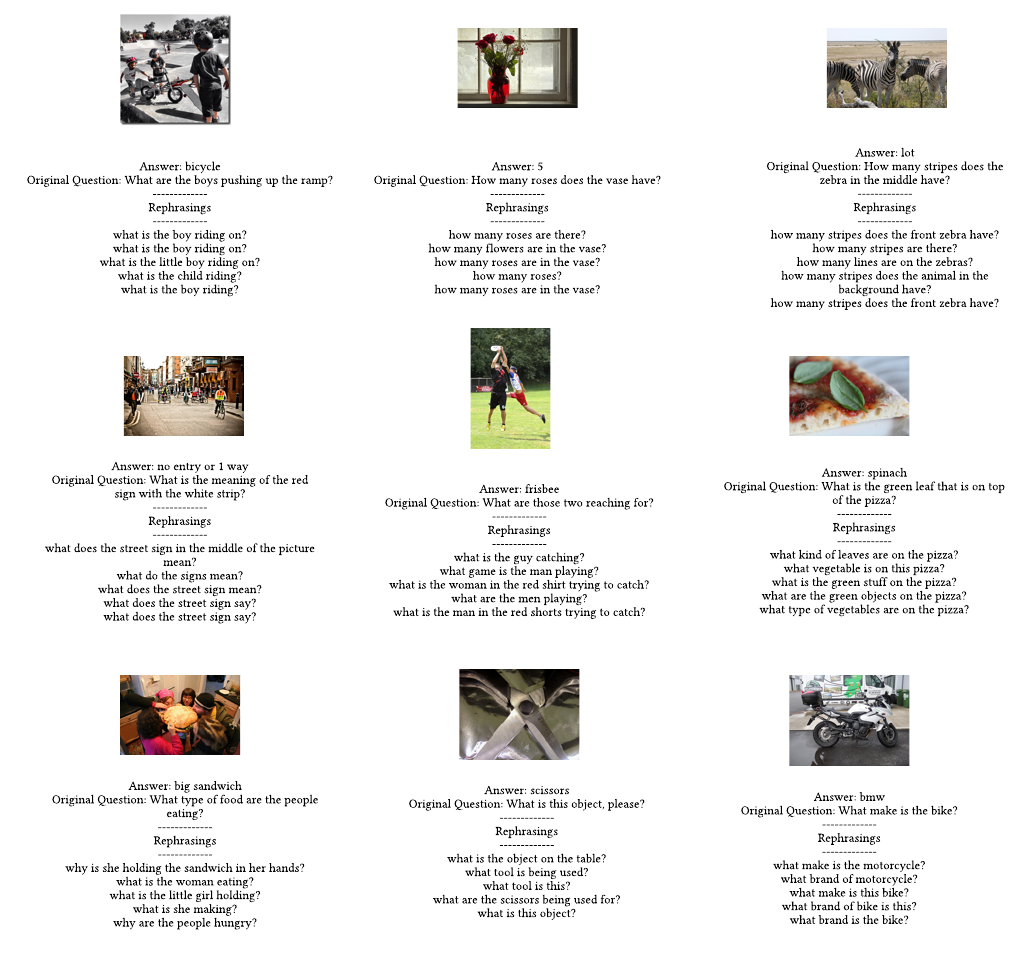}
    \caption{More examples of generated rephrasings.}
    \label{fig:extended_qualitative}
\end{figure}
 \fi

\end{document}